\documentclass{article}

\usepackage[main,final]{neurips_2026}

\usepackage[utf8]{inputenc} % allow utf-8 input
\usepackage[T1]{fontenc}    % use 8-bit T1 fonts
\usepackage{hyperref}       % hyperlinks
\usepackage{url}            % simple URL typesetting
\usepackage{booktabs}       % professional-quality tables
\usepackage{amsfonts}       % blackboard math symbols
\usepackage{nicefrac}       % compact symbols for 1/2, etc.
\usepackage{microtype}      % microtypography
\usepackage{xcolor}         % colors
\usepackage{tabularx}
\usepackage{multirow}
\usepackage{array}
\usepackage{enumitem}
\usepackage{xcolor}
\usepackage{graphicx}
\usepackage{tikz}
\usepackage{amsmath}
\usepackage{tabularx}
\usepackage{listings}
\usepackage{array}

\title{Building Trustworthy Graph-Agentic RAG for Social Good: Architectures, Failure Propagation, and Assurance by Construction}

\author{%
Vijay Bommireddy \\
Independent Researcher \\
\texttt{vijaybommireddy1@gmail.com} \\
\And
Raviteja Bommireddy \\
IIITDM Kancheepuram \\
\texttt{ravitejab1978@gmail.com}
}

\begin{document}

\maketitle

\begin{abstract}
Graph-agentic retrieval-augmented generation combines structured evidence with adaptive controllers that can plan retrieval, traverse relations, verify intermediate claims, delegate subtasks, or use tools. These systems are relevant when answers depend on relations across documents, entities, time, or institutions, but the same combination creates coupled failure paths: a graph-construction defect can become retrieved evidence, alter later control decisions, and propagate toward a consequential outcome. We examine how such systems can be designed and evaluated for social-good settings in which freshness, authorization, traceability, oversight, and recourse matter alongside answer quality. The survey is organized by graph substrate, graph lifecycle, agent function, coordination pattern, and authority boundary; it distinguishes graph-based retrieval from observation-dependent graph control and synthesizes the resulting risks as an evidence-to-action failure chain. Our central contribution is an assurance-by-construction blueprint built around five interface contracts. The contracts make provenance, temporal validity, authorization, uncertainty, and recoverability explicit and checkable as information moves from source ingestion to human action. A worked public-benefit information case shows how the blueprint affects choices about graph design, agent permissions, verification, abstention, and operating authority. Rather than pool results from heterogeneous benchmarks, we specify evaluation evidence at the graph, trajectory, claim, coordination, and outcome levels. The resulting reference architecture makes explicit what should remain observable, checkable, and controllable when graph-agentic RAG is considered for consequential applications.
\end{abstract}

\section{Introduction}
\label{sec:introduction}

Retrieval-augmented generation (RAG) grounds a language model in external
evidence rather than relying only on parametric memory
\citep{lewis2020rag,karpukhin2020dpr}. Graph-based RAG adds structure: nodes can
represent entities, passages, events, or communities, while edges can represent
typed relations, document links, temporal order, or shared context
\citep{edge2024graphrag,he2024gretriever,mavromatis2025gnnrag}. Agentic control
adds a different capability: a system can revise a query, choose a tool,
traverse another path, verify an intermediate claim, or delegate a subtask in
response to what it observes \citep{yao2023react,asai2024selfrag,
jeong2024adaptiverag}. We use \emph{graph-agentic RAG} for systems in which such
adaptive control governs at least one material graph operation.

Graph structure and adaptive control are useful when a task depends on relations across institutions, rules, time periods, or multiple pieces of evidence. They also create coupled failure paths. A stale or unsupported graph assertion can be retrieved, change a controller's next step, pass through a handoff, and appear downstream as a confident answer. Retrieval poisoning, graph poisoning, poisoned agent memory, and indirect prompt injection show that such propagation is not merely hypothetical \citep{zou2025poisonedrag,liang2026graphragfire,
chen2024agentpoison,zhan2024injecagent,debenedetti2024agentdojo}.

For social-good applications, the evaluation target cannot stop at answer accuracy. Information systems can affect access to care or public benefits, crisis guidance, and the direction of scientific work. Their evidence may be fragmented, rapidly changing, sensitive, or governed by different authorities.
The questions therefore include: Was the source authorized and current? Which source supports each node, edge, and claim? Did an agent exceed the user's authority? Could a competent reviewer intercept the error? Can an affected person contest the outcome? These are deployment questions about institutions and consequences, not only model questions \citep{tomasev2020aiforsocialgood,
tabassi2023airmf,worldhealthorganization2021ethics,
lyons2021contestability}.

Table \ref{tab:related} positions this work relative to representative GraphRAG, agentic RAG,
and graph-augmented agents, as well as empirical studies of
when graphs or agentic retrieval help \citep{peng2025graphragsurvey,
singh2025agenticragsurvey,liu2026graphagents,
xiang2026whengraphs,ferrazzi2026agenticworth}. Existing work primarily explains how these systems are built, what tasks they solve, and when added graph or agent mechanisms improve utility. Our organizing question is different: what must remain valid, observable, and controllable as evidence and authority move from source ingestion toward consequential action? Within the reviewed corpus, we did not identify a survey organized jointly around cross-layer failure propagation, checkable interface obligations, and their application to a social-good design.

% \begin{table}[t]
% \centering
% \caption{Positioning against the closest survey and comparison literature. The
% last column identifies the question this survey takes as primary.}
% \label{tab:related}
% \footnotesize
% \setlength{\tabcolsep}{3pt}
% \renewcommand{\arraystretch}{1.08}
% \begin{tabularx}{\linewidth}{@{}p{0.20\linewidth}YY@{}}
% \toprule
% Work & Main organizing question & Gap for consequential deployment \\
% \midrule
% GraphRAG surveys \\ \citep{peng2025graphragsurvey,zhu2026graphfunctionalities}
% & How are graphs built, retrieved, and evaluated?
% & Do not follow evidence and authority through adaptive control to human outcome. \\
% Agentic RAG surveys \\ \citep{singh2025agenticragsurvey,deng2026datacentric}
% & How do agents plan, retrieve, \\ reflect, and manage data?
% & Graph-specific integrity and traversal \\ failures are not the organizing unit. \\
% Graph--agent surveys \\ \citep{liu2026graphagents,chen2026agenticgraphrag}
% & How do graphs support agents \\ and agentic GraphRAG systems?
% & Limited treatment of checkable interface \\ obligations, operating authority, and recourse. \\
% Choice studies \\ \citep{xiang2026whengraphs,ferrazzi2026agenticworth}
% & When do graph or agentic variants improve \\ measured utility and cost?
% & Benchmark utility does not establish \\ deployability under consequential harm. \\
% This survey
% & What must remain valid, observable, \\ and controllable from source to outcome?
% & Connects mechanisms, propagating failures, \\ interface contracts, and a worked social-good design. \\
% \bottomrule
% \end{tabularx}
% \end{table}

\begin{table}[t]
\centering
\caption{Positioning relative to representative survey and comparison literature. The final column identifies the deployment question emphasized in this survey.}
\label{tab:related}
\footnotesize
\setlength{\tabcolsep}{3pt}
\renewcommand{\arraystretch}{1.08}

\begin{tabularx}{\linewidth}{@{}p{0.18\linewidth}X X@{}}
\toprule
Work & Main organizing question & Gap for consequential deployment \\
\midrule
GraphRAG surveys 
% \citep{peng2025graphragsurvey,zhu2026graphfunctionalities}
& How are graphs built, retrieved, and evaluated?
& Do not follow evidence and authority through adaptive control to human outcome. \\
Agentic RAG surveys 
% \citep{singh2025agenticragsurvey,deng2026datacentric}
& How do agents plan, retrieve, reflect, and manage data?
& Graph-specific integrity and traversal failures are not the organizing unit. \\
Graph--agent surveys 
% \citep{liu2026graphagents,chen2026agenticgraphrag}
& How do graphs support agents and agentic GraphRAG systems?
& Interface-level assurance obligations, operating authority, and recourse are not the primary organizing focus. \\
Choice studies 
% \citep{xiang2026whengraphs,ferrazzi2026agenticworth}
& When do graph or agentic variants improve measured utility and cost?
& Benchmark utility alone does not establish suitability for consequential deployment \\
\textbf{This survey}
& What must remain valid, observable, and controllable from source to outcome?
& Connects mechanisms, propagating failures, interface contracts, and a worked social-good design. \\
\bottomrule
\end{tabularx}
\end{table}

\paragraph{Research question.}
\emph{How should graph-agentic RAG systems be designed, evaluated, and bounded so that relational evidence remains traceable and operating authority remains controllable from source to consequential outcome?}

\paragraph{Contributions.}
We answer this question through four connected contributions:
\begin{enumerate}
    \item a mechanism-centered synthesis of graph substrates, lifecycles, adaptive control, coordination, and authority boundaries.
    \item an evidence-to-action failure model that explains how defects in graph construction, retrieval, reasoning, coordination, and oversight propagate rather than occur as independent buckets
    \item an \emph{assurance-by-construction blueprint} built from five interface contracts evidence, retrieval, reasoning, delegation, and outcome with preventive, detective, and recovery responsibilities
    \item a worked public-benefit information design plus a reporting agenda for evaluating graph quality, trajectories, grounding, coordination, abstention, human review, and recourse.
\end{enumerate}

The blueprint is a literature-grounded design synthesis rather than a certification procedure or safety guarantee; it is intended to make missing evidence, broken obligations, and excessive operating authority explicit and testable.

\section{Scope and evidence discipline}
\label{sec:method}

We reviewed work published or posted from January 2020 through 29 August 2026, with earlier foundational studies added through citation chaining when directly relevant to governance or human oversight. Searches combined terms for RAG, graphs, knowledge graphs, agents, planning, and orchestration with provenance, freshness, robustness, poisoning, privacy, fairness, oversight, and contestability. We included systems with a graph or retrieval substrate, generation component, and enough control detail to identify adaptive decisions, together with assurance work directly relevant to the resulting failure surfaces. We excluded visualization-only graphs, graph-embedding work without retrieval and generation, duplicate reports, and fixed pipelines labelled as agents. Primary proceedings and publisher records were used to verify bibliographic identity and publication status; an evidence ledger records the claims supported by each source. Because tasks, graphs, models, metrics, and hardware differ substantially, we compare reported mechanisms and evaluation scope rather than pool scores or rank architectures.

\section{From Graph Retrieval to Graph-Agentic Systems}
\label{sec:landscape}

The term \emph{GraphRAG} covers materially different systems. Some organize a
corpus into entity and community summaries; others retrieve a subgraph from an
existing knowledge graph, use a graph neural network to identify paths, or
construct relations at query time. Likewise, \emph{agentic} can denote anything
from a fixed sequence with an LLM in the loop to a controller that observes an
intermediate result and changes its next action. Treating both labels as binary
properties obscures the engineering decisions that create benefit and risk.

We therefore use an operational boundary: a system is \emph{graph-agentic} when
(i) nodes, edges, paths, or graph summaries are material evidence for the
answer, and (ii) an observation made during execution can change a subsequent
graph operation, retrieval, verification step, delegation, or tool call. This
definition excludes graphs used only for visualization and fixed pipelines
whose control flow cannot respond to evidence. It also avoids presenting flat
RAG, GraphRAG, and agentic RAG as rungs on a maturity ladder. Each can be the
right design for a different information problem.

\subsection{Five dimensions that determine system behavior}

Table \ref{tab:design-space} organizes systems by mechanisms rather than product
names. The first two dimensions describe the evidence substrate: what a graph
asserts and how it changes. The next two describe control: what an agent may
decide and how those decisions move among components. The final dimension
describes authority: whether the output informs a person or can change the
world. That last distinction is essential in social-good applications. The same
retrieval path can be acceptable for exploratory search and unacceptable as
the sole basis for denying a service.

\begin{table}[t]
\centering
\caption{A mechanism-centered design space for graph-agentic RAG. The
dimensions are deliberately non-exclusive: a deployed system may occupy
several cells in each row.}
\label{tab:design-space}
\footnotesize
\setlength{\tabcolsep}{3pt}
\renewcommand{\arraystretch}{1.08}
\begin{tabularx}{\linewidth}{@{}p{0.18\linewidth}X X@{}}
\toprule
Dimension & Representative choices & Assurance consequence \\
\midrule
Graph substrate
& document/entity graph; knowledge graph; heterogeneous or temporal graph
& Defines what an edge asserts and which provenance must be retained. \\
Graph lifecycle
& curated; extracted offline; updated incrementally; assembled at query time
& Determines validation, versioning, deletion, and freshness duties. \\
Agent function
& query planning; traversal; evidence critique; synthesis; tool use
& Determines which intermediate choices require traces and stopping rules. \\
Coordination
& single controller; specialist agents; planner--executor; debate or verifier
& Creates handoffs at which errors, instructions, and permissions can propagate. \\
Authority boundary
& information only; recommendation; draft action; authorized execution
& Sets review, escalation, authentication, logging, and recourse requirements. \\
\bottomrule
\end{tabularx}
\end{table}

Graph construction and agent control interact. An extracted edge can compress
an uncertain sentence into an apparently definite relation; a traversal policy
can then repeatedly select that edge; a verifier that receives only the
retrieved subgraph may be unable to see the original qualification. Conversely,
a carefully curated graph does not make an agent safe if it can ignore access
rules, call an unauthorized tool, or convert advice into action. Assurance must
therefore attach to interfaces, not merely to model components.

\paragraph{Minimal-design guide.}
Start with flat or hybrid RAG when the answer lies in a bounded, current
document set. Add a graph when multi-hop, entity, community, or temporal
relations are material; deterministic graph queries may still suffice. Add
agentic control when an intermediate observation must change the next search,
verification, or tool step. Combine both only when the task needs relational
evidence \emph{and} observation-dependent graph operations, and only when the
resulting traces, stopping rules, and permissions can be enforced. This is a
design sequence, not a claim that one family is more advanced. Operating
authority is assessed separately in Section \ref{sec:blueprint}.

\subsection{What representative systems establish}

Table \ref{tab:systems} samples influential and recent systems to show the range of
graph roles and adaptive control. It is not a leaderboard. GraphRAG demonstrates
global, community-oriented summarization; G-Retriever and GNN-RAG demonstrate
learned subgraph or path retrieval; Think-on-Graph makes iterative graph
traversal explicit; HybGRAG combines relational and textual evidence; and GeAR
and Graph-R1 move closer to adaptive graph operation. These studies establish that graph structure and iterative control can be useful on suitable tasks 
\citep{edge2024graphrag,he2024gretriever,sun2024thinkongraph,
mavromatis2025gnnrag,shen2025gear}. They do not, by themselves, establish that
an architecture is safe under stale policies, adversarial evidence, sensitive
attributes, cross-agent handoffs, or consequential operating authority.

\begin{table}[t]
\centering
\caption{Representative systems illustrating distinct graph and control
mechanisms. Reported results are not pooled because their tasks and protocols
differ. ``Adaptive'' means that an intermediate observation can change a later
retrieval or reasoning step.}
\label{tab:systems}
\scriptsize
\setlength{\tabcolsep}{2.5pt}
\renewcommand{\arraystretch}{1.07}
\begin{tabularx}{\linewidth}{@{}p{0.19\linewidth}X X X@{}}
\toprule
System & Material graph role & Adaptive control & Evaluation boundary \\
\midrule
GraphRAG \citep{edge2024graphrag}
& entity/relation graph and community summaries
& query-focused use of precomputed summaries; not an acting agent
& corpus-level question-focused summarization \\
G-Retriever \citep{he2024gretriever}
& retrieves a compact subgraph for graph QA
& learned retrieval and generation pipeline; no delegated action
& textual-graph understanding and QA \\
Think-on-Graph \citep{sun2024thinkongraph}
& knowledge-graph paths are the reasoning substrate
& iteratively selects relations and entities from observations
& knowledge-graph reasoning and QA \\
HybGRAG \citep{lee2025hybgrag}
& jointly uses textual and relational evidence
& hybrid retrieval, without autonomous external action
& QA over text-rich relational knowledge bases \\
GNN-RAG \citep{mavromatis2025gnnrag}
& a GNN retrieves reasoning paths for an LLM
& graph retrieval is learned; agentic replanning is not central
& multi-hop knowledge-graph QA \\
GeAR \citep{shen2025gear}
& graph-enhanced retrieval supplies structured evidence
& agent adaptively refines retrieval and generation
& multi-hop and complex QA \\
Graph-R1 \citep{luo2025graphr1}
& graph operations are exposed to an agentic policy
& end-to-end policy learns graph retrieval actions
& preprint evaluations on graph-reasoning tasks \\
\bottomrule
\end{tabularx}
\end{table}

The appropriate design question is consequently not ``Which RAG is most
advanced?'' It is: \emph{Which relations must be represented, which decisions
must adapt at run time, and what evidence and authority must cross each
interface?} The remainder of the survey follows those interfaces from source to
outcome.

\section{The Evidence-to-Action Failure Chain}
\label{sec:failures}

\begin{figure}
  \centering
  \includegraphics[width=\linewidth]{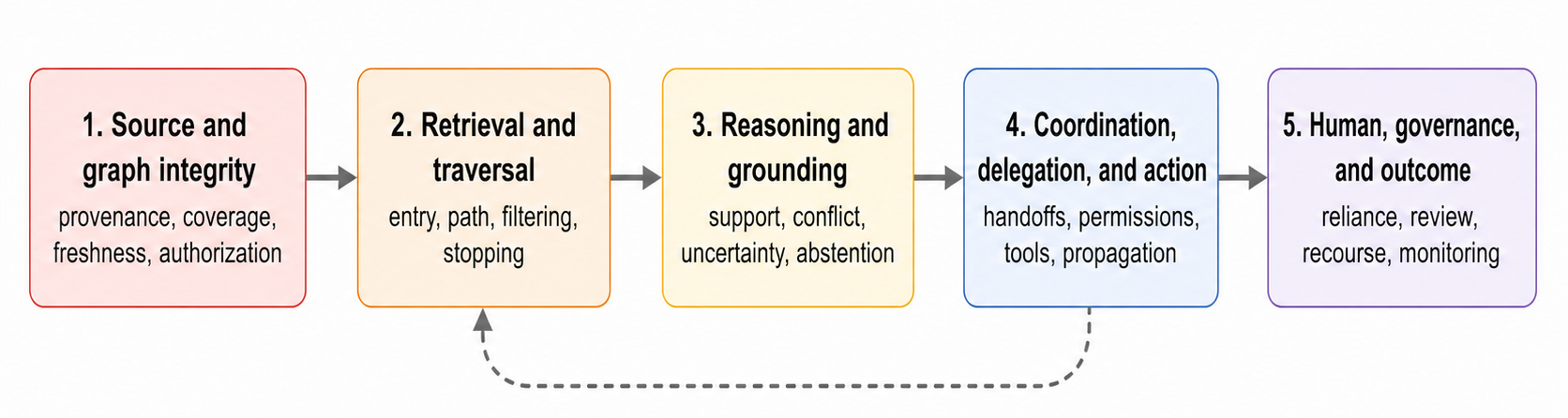}
  \caption{Five stages at which a defect can originate, be amplified, or be intercepted. The feedback path represents adaptive retrieval after a reasoning or coordination observation.}
  \label{fig:failure-chain}
\end{figure}

Component taxonomies make failures look separable: graph errors belong to data
engineering, hallucinations to the language model, and automation errors to an
agent. In a graph-agentic system, however, the output of one component becomes
the evidence or control state of the next. We model this as the five-stage
chain in Figure \ref{fig:failure-chain}. The stages are non-exclusive and can repeat;
an agent may return to retrieval, modify memory, or delegate to another agent.

\paragraph{1. Source and graph integrity.}
A source can be unofficial, stale, incomplete, malicious, or unauthorized for
the user. Graph construction adds transformation risk: entity resolution can
merge distinct people, relation extraction can remove qualification, and an
edge can survive after its source is corrected. These defects are consequential
because graph traversal treats structure as evidence. Retrieval-store and
graph-specific poisoning studies demonstrate that a small number of inserted
records can redirect later generation \citep{zou2025poisonedrag,
zhao2026knowledgepoisoning,liang2026graphragfire}. Privacy and fairness are also
properties of the assembled evidence, not only the generator
\citep{zeng2024ragprivacy,wu2025ragfairness}.

\paragraph{2. Retrieval and traversal.}
Even a valid graph can yield an invalid evidence set. Entry-node selection can
miss the relevant entity; popularity-biased traversal can overrepresent a
well-connected institution; approximate search can omit an exception; and a
poor stopping rule can turn one irrelevant hop into an entire reasoning path.
Temporal questions are especially fragile when retrieval does not respect the
validity interval of facts \citep{vu2024freshllms,qian2024timer4}. Adaptive
retrieval makes these errors path-dependent: the first returned edge influences
the next query and therefore the evidence that a later verifier can see.

\paragraph{3. Reasoning and grounding.}
Retrieved evidence may be relevant without entailing a claim. The generator can
combine incompatible versions, infer a relation absent from the source, or
express uncertainty as a definitive answer. Citation correctness and answer
quality must therefore be measured separately \citep{gao2023alce,
es2024ragas,saadfalcon2024ares}. A graph path is not automatically a proof: each
edge needs semantic meaning, source support, and time and access context.

\paragraph{4. Coordination, delegation, and action.}
An agent may turn a reasoning error into a new retrieval, a memory entry, a
message to another agent, or a tool call. Indirect instructions in retrieved
content can alter the control path, while poisoned memory can trigger targeted
behavior \citep{zhan2024injecagent,debenedetti2024agentdojo,
chen2024agentpoison}. Multi-agent systems add propagation through ambiguous
handoffs and communication topologies \citep{cemri2025masfailures,
shen2025propagation}. Delegation is also an authorization problem: an agent
should not acquire rights merely because another agent asked it to act
\citep{south2025delegation}.

\paragraph{5. Human, governance, and outcome.}
A technically faithful answer can still cause harm if it is shown without an
effective warning, routed to the wrong decision-maker, or treated as a binding
decision. Nominal ``human review'' is insufficient when reviewers lack time,
source access, authority to disagree, or a path for correction
\citep{green2019algorithmloop,raji2020auditing}. Outcome assurance must include
monitoring, incident ownership, appeal, and correction of downstream state
\citep{lyons2021contestability,tabassi2023airmf}.

\subsection{Why propagation changes evaluation}

Three mechanisms make end-to-end accuracy an incomplete safety signal.
\emph{Amplification} occurs when a single bad edge influences many paths or
agents. \emph{Relabeling} occurs when an upstream provenance defect appears
downstream as a reasoning or user error. \emph{Observability loss} occurs when
summaries, handoffs, or actions omit the trace needed to diagnose the origin.
A trustworthy system must therefore measure not only whether a failure occurs,
but where it originates, whether it is detected before the next interface, how
far it propagates, and whether its effects can be reversed.

\section{Assurance by Construction}
\label{sec:blueprint}

The failure chain suggests a constructive requirement: evidence, claims,
instructions, and action requests should not cross a system boundary without
enough context to validate them. We express this requirement through five
\emph{interface contracts}. For contract $C_i$,
\begin{equation}
  C_i = (I_i, P_i, O_i, V_i, F_i),
  \label{eq:contract}
\end{equation}
where $I_i$ denotes the accepted input, $P_i$ the governing policy, $O_i$ the
observable artifact emitted by the stage, $V_i$ the validation performed before
release, and $F_i$ the response to validation failure. Unlike a component
checklist, this formulation places obligations at system boundaries: downstream
stages can reject incomplete artifacts, and failures can be traced to the first
violated contract.

\begin{figure}
  \centering
  \includegraphics[width=\linewidth]{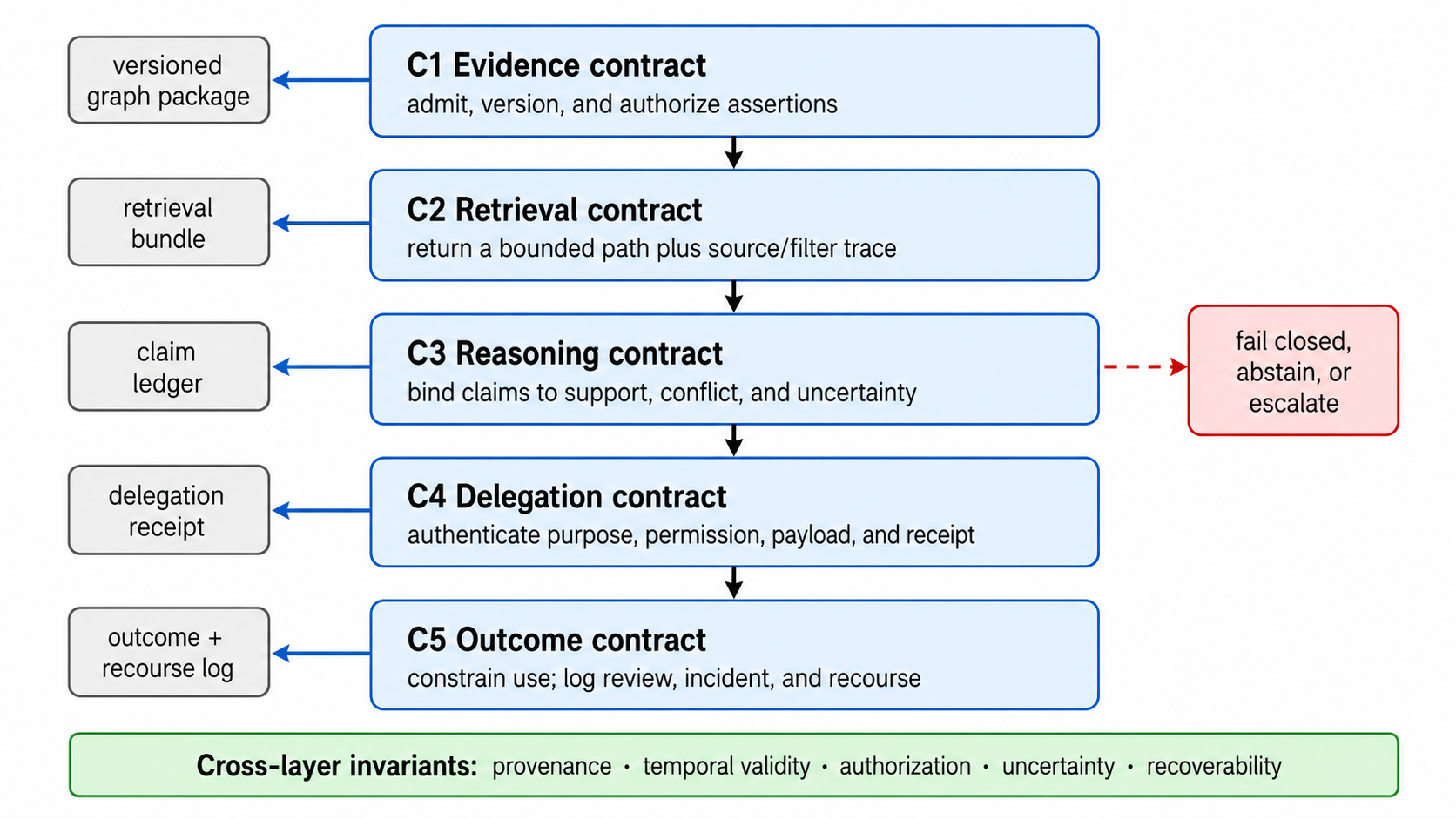}
  \caption{Assurance-by-construction blueprint. Each interface emits a
  checkable artifact before evidence or authority is propagated. Failed checks
  may block, abstain, or escalate; cross-layer invariants remain visible across
  transformations.}
  \label{fig:blueprint}
\end{figure}

\subsection{Five contracts across the system boundary}

Table~\ref{tab:contracts} specifies the artifact, validation, and failure
response associated with each interface. Exact schemas and thresholds are
domain-specific; the recurring requirement is that evidence and authority do
not cross a boundary without an observable basis for checking them.

\begin{table}[t]
\centering
\caption{Proposed interface contracts for assurance by construction. Each
contract emits a checkable artifact, applies validation before release, and
defines a fail-closed response.}
\label{tab:contracts}
\footnotesize
\setlength{\tabcolsep}{2.5pt}
\renewcommand{\arraystretch}{1.08}
\begin{tabularx}{\linewidth}{@{}p{0.19\linewidth}X X@{}}
\toprule
Contract & Required artifact and validation & Failure response \\
\midrule

C1 Evidence
& versioned node/edge record with source span, issuer, validity interval,
extraction method, and access label; check provenance, schema, freshness, and
authorization
& quarantine assertion; retain prior valid version; trigger source review \\

C2 Retrieval
& query, graph version, entry nodes, traversed path, filters, scores, and
stopping reason; check path legality, access constraints, and
coverage
& retry within bounds; broaden evidence; ask for missing context; abstain if
coverage remains insufficient \\

C3 Reasoning
& claim ledger links each material to supporting or conflicting
evidence, time, status, entailment, citation support, and temporal
consistency
& revise or qualify claim; expose conflict; abstain or request review \\

C4 delegation \& Capability
& authenticated purpose, scoped capability, payload provenance, expiry, and
receipt; check least privilege and instruction--data separation
& deny tool call or handoff; revoke capability; contain affected state \\

C5 Outcome
& intended use, authority level, required review, override record,
and correction or recourse log; check permitted use and review completion
& block or pause automation; route to accountable owner; correct and monitor \\

\bottomrule
\end{tabularx}
\end{table}

The contracts impose three broader design principles. First, graph assertions
remain tied to their provenance, version, temporal scope, and access conditions
rather than becoming anonymous evidence. Second, retrieval and reasoning retain
enough context to inspect how a path or material claim was produced, including
conflicting evidence and applicability conditions. Third, authority does not
propagate implicitly: a retrieved instruction, model output, or inter-agent
message cannot create new tool or data permissions. This last requirement is
particularly important under indirect prompt injection and poisoned evidence
\citep{zhan2024injecagent,debenedetti2024agentdojo,south2025delegation}.
The outcome contract extends the same principle to deployment by binding system
output to an intended use, accountable review, and a route for correction.

\subsection{Invariants and operating authority}

Five invariants span the contracts: \emph{provenance}, \emph{temporal validity},
\emph{authorization}, \emph{uncertainty}, and \emph{recoverability}. A
transformation violates the proposed assurance conditions when it discards
information required to preserve one of these invariants, even if the final
answer is correct.

Architecture and operating authority should therefore be chosen separately.
Adaptive graph retrieval may be appropriate while automated final action
remains prohibited; conversely, even a simple system can be unsafe when its
output is treated as binding. Appropriate operating modes range from
exploratory information and monitored assistance to mandatory pre-action review
or no automated final action.

\paragraph{When not to build graph-agentic RAG.}
Prefer a simpler design when authoritative evidence is bounded, required
relations can be handled deterministically, adaptive control is unnecessary, or
the required trace cannot be maintained. Consequential decisions should remain
human-led when source legitimacy, affected-population risk, or effective
correction and recourse cannot be established.
\section{Illustrative design: Public-Benefit Information}
\label{sec:case}

We now apply the blueprint to one consequential but common information task.
The purpose is to demonstrate the decisions and artifacts produced by the
synthesis, not to report a deployment or claim measured effectiveness.

\paragraph{Scenario.}
A resident asks: ``I moved to a different county, my caregiving income changed,
and one page mentions an exception. Which programs may be relevant, which rule
versions apply, and what information is still missing?'' The evidence spans
program rules, jurisdiction, effective dates, definitions, exceptions, and
issuing offices. A single semantic search can retrieve a plausible general rule
while omitting the exception or using a page from the wrong date.

\paragraph{Architecture decision.}
The relational and temporal structure is material, so the system uses a typed,
versioned graph. The path also depends on observations: after identifying the
county and date, the controller may need to ask a question, inspect an
exception, or compare two active rule versions. A bounded graph-agentic design
is therefore justified for information retrieval. A multi-agent architecture
and execution tools are not justified: they add handoffs and authority without
a task requirement. If the service covered only one current, bounded policy document, the
minimal-design guide in Section~\ref{sec:landscape} would instead favor
ordinary RAG. Table~\ref{tab:worked-case} shows how the five contracts
instantiate this design and identifies the conditions under which the system
must ask for missing information, abstain, deny an action, or defer to an
authorized decision-maker.

\begin{table}[t]
\centering
\caption{Applying the contracts to a public-benefit information assistant.}
\label{tab:worked-case}
\scriptsize
\setlength{\tabcolsep}{2.5pt}
\renewcommand{\arraystretch}{1.08}
\begin{tabularx}{\linewidth}{@{}p{0.17\linewidth}X X@{}}
\toprule
Contract & Concrete design decision & Observable stop condition \\
\midrule
C1 Evidence
& versioned graph links program, jurisdiction, rule, exception, effective date, required document, and issuing office to official source spans
& unofficial or expired rule is quarantined; unsupported edge is not traversable \\
C2 Retrieval
& controller first resolves jurisdiction and date, then follows eligibility and exception paths under hop and source-tier limits
& missing jurisdiction, unresolved identity, or incomplete exception coverage triggers a question or abstention \\
C3 Reasoning
& claim ledger separates possible relevance, missing evidence, conflict, and authoritative determination
& no claim is emitted if its rule version or source span is absent \\
C4 Delegation \& Capability
& no multi-agent handoff is needed; a bounded controller may search approved sources but cannot submit, enroll, or disclose records
& instruction embedded in retrieved text cannot expand tool or data permissions \\
C5 Outcome
& The output provides cited information that users can discuss with a caseworker, along with options to request corrections or appeal decisions.
& final eligibility or adverse action remains outside the system \\
\bottomrule
\end{tabularx}
\end{table}

\paragraph{A complete trace.}
At ingestion, C1 retains the source, issuer, jurisdiction, validity interval,
and version for each rule and exception, quarantining unsupported or unofficial
evidence. At query time, C2 records the applicable jurisdiction and date,
retrieval path, filters, and stopping reason; missing information triggers a
clarifying question rather than a guess. C3 maps each material claim to its
supporting evidence and keeps conflicting rule versions visible. C4 restricts
the controller to approved sources and prevents retrieved instructions from
expanding its permissions. C5 presents the result as cited information and
routes any final determination to an authorized caseworker, while preserving
channels for correction or contestation.

\paragraph{What the example changes.}
The blueprint changes the design objective from answering a multi-hop query to
controlling how evidence and authority move through the system. Graph assertions
become source-backed and time-bounded, retrieval paths remain inspectable, the
controller operates with least privilege, and consequential decisions remain
outside its authority. The system therefore gains relational and adaptive
retrieval without conflating technical capability with institutional authority.
\section{Evaluation and Reporting Agenda}
\label{sec:evaluation}

Existing RAG evaluation usefully separates retrieval relevance, answer
faithfulness, and citation quality \citep{gao2023alce,es2024ragas,
saadfalcon2024ares}. Graph-agentic systems require these measures, but also need
evidence about graph assertions, execution trajectories, delegated authority,
and human effects. Table \ref{tab:evaluation} defines a minimum evaluation
specification aligned with the five contracts.

\begin{table}[t]
\centering
\caption{Minimum evaluation evidence for a consequential graph-agentic RAG
system. No single metric is sufficient.}
\label{tab:evaluation}
\scriptsize
\setlength{\tabcolsep}{2.5pt}
\renewcommand{\arraystretch}{1.08}
\begin{tabularx}{\linewidth}{@{}p{0.19\linewidth}X X@{}}
\toprule
Level & Measure under ordinary and stressed conditions & Report with result \\
\midrule
Graph
& assertion support, source coverage, temporal validity, duplicate/entity-resolution error, access-policy leakage
& graph version, construction method, sampled audit protocol, update/deletion behavior \\
Trajectory
& relevant-path coverage, irrelevant hops, search cost, stopping accuracy, path stability, replay success
& query/tool trace, budgets, seeds, filters, retry and abstention behavior \\
Claim
& claim support and completeness, citation correctness, contradiction handling, uncertainty and abstention calibration
& claim ledger and verifier inputs, including original source spans \\
Coordination
& permission violations, instruction/data confusion, handoff loss, injected-error detection, propagation and containment
& agent topology, capabilities, messages, tool calls, denied actions, affected state \\
Outcome
& reviewer correction, override, reliance, subgroup error, time-to-remedy, appeal access, downstream correction
& intended use, decision owner, incident protocol, monitoring window, affected population \\
\bottomrule
\end{tabularx}
\end{table}

\subsection{Evaluate propagation, not only final error}

For a documented set of naturally occurring, counterfactual, or deliberately
injected defects, evaluation should record four quantities. \emph{Origin
detection} asks whether the initiating defect is recognized at its own stage.
\emph{Propagation depth} is the number of interfaces crossed before detection.
\emph{Containment} is the fraction of triggered defects prevented from crossing
the system's authorized outcome boundary. \emph{Recovery completeness} is the
fraction of affected graph, memory, logs, outputs, and human records that can be
identified and corrected. These quantities reveal systems that answer ordinary
questions well but fail dangerously under stale, conflicting, poisoned, or
permission-mismatched evidence.

Stress tests should vary graph perturbations, source freshness, ambiguous
entities, missing exceptions, conflicting authorities, indirect instructions,
agent handoffs, and tool permissions. Each perturbation needs an expected
contract response: continue, seek more evidence, ask the user, abstain, deny an
action, or escalate. This supports falsifiable testing without assuming that
one benchmark score represents deployment readiness.

\subsection{Compare against the simplest adequate design}

Every graph-agentic evaluation should include a non-agentic and, where
appropriate, non-graph baseline under the same corpus, model, context budget,
and outcome definition. Report quality together with latency, token and tool
cost, trace completeness, and intervention burden. The relevant result is not
whether the most elaborate architecture wins somewhere, but whether its added
mechanism yields a material benefit that compensates for additional failure
surface and operating cost \citep{xiang2026whengraphs,
ferrazzi2026agenticworth}.

\subsection{Research priorities}

Four gaps are especially actionable. First, graph benchmarks need assertion-
level provenance, time validity, access labels, and controlled corruption so
that source and traversal failures can be separated. Second, trajectory
benchmarks need task-equivalent alternatives and expert-auditable stopping
points, not only final answers. Third, multi-agent evaluations need permission
and handoff annotations that expose error propagation across roles. Fourth,
social-good studies need outcome protocols co-designed with domain workers and
affected communities, including burden, subgroup effects, contestability, and
correction after release. Such studies would prospectively test the contracts
proposed here rather than treating them as established guarantees.

\section{Limitations}
\label{sec:limitations}

Screening and synthesis used one primary reviewer with a separate consistency
audit, not independent dual screening or inter-rater agreement. The search
favors English-language and indexed literature and may miss unpublished
operational experience. Several fast-moving families are represented by
preprints whose status can change. Heterogeneous tasks, graphs, models, and
metrics prevent defensible score pooling, so we compare mechanisms rather than
rank architectures. Finally, the contracts and worked case are a
literature-grounded synthesis, not a prospectively validated standard,
deployed service, or legal determination. Their utility and burden require
testing with domain practitioners and affected communities. These limits bound
our claims, the artifacts in sections \ref{sec:blueprint} \& \ref{sec:evaluation} make the missing evidence concrete enough to test.

\section{Conclusion}
\label{sec:conclusion}

Graph-agentic RAG is useful when relational evidence and observation-dependent
control are both material; it is not a universal upgrade. In consequential
settings, evidence quality and authority must survive the complete path from
source to outcome. The proposed failure chain and five contracts make that path
observable through versioned, replayable, and recoverable artifacts. They are
an evaluable reference architecture, not a guarantee. Future graph audits,
trajectory stress tests, permission-aware agent evaluation, and outcome studies
should test whether teams can locate a broken obligation, contain propagation,
and correct technical and human state. Until then, the safest design is the
least complex one whose evidence is traceable, authority bounded, and effects
reversible.

\bibliographystyle{plainnat}
\bibliography{references}

@inproceedings{lewis2020rag,
  title     = {Retrieval-Augmented Generation for Knowledge-Intensive {NLP} Tasks},
  author    = {Lewis, Patrick and Perez, Ethan and Piktus, Aleksandra and Petroni, Fabio and Karpukhin, Vladimir and Goyal, Naman and K{\"u}ttler, Heinrich and Lewis, Mike and Yih, Wen-tau and Rockt{\"a}schel, Tim and Riedel, Sebastian and Kiela, Douwe},
  booktitle = {Advances in Neural Information Processing Systems},
  volume    = {33},
  pages     = {9459--9474},
  year      = {2020}
}

@inproceedings{karpukhin2020dpr,
  title     = {Dense Passage Retrieval for Open-Domain Question Answering},
  author    = {Karpukhin, Vladimir and Oguz, Barlas and Min, Sewon and Lewis, Patrick and Wu, Ledell and Edunov, Sergey and Chen, Danqi and Yih, Wen-tau},
  booktitle = {Proceedings of EMNLP},
  pages     = {6769--6781},
  year      = {2020},
  doi       = {10.18653/v1/2020.emnlp-main.550}
}

@inproceedings{yao2023react,
  title     = {{ReAct}: Synergizing Reasoning and Acting in Language Models},
  author    = {Yao, Shunyu and Zhao, Jeffrey and Yu, Dian and Du, Nan and Shafran, Izhak and Narasimhan, Karthik R. and Cao, Yuan},
  booktitle = {International Conference on Learning Representations},
  year      = {2023},
  url       = {https://openreview.net/forum?id=WE_vluYUL-X}
}

@inproceedings{asai2024selfrag,
  title     = {{Self-RAG}: Learning to Retrieve, Generate, and Critique through Self-Reflection},
  author    = {Asai, Akari and Wu, Zeqi and Wang, Yizhong and Sil, Avirup and Hajishirzi, Hannaneh},
  booktitle = {International Conference on Learning Representations},
  year      = {2024},
  url       = {https://openreview.net/forum?id=hSyW5go0v8}
}

@inproceedings{jeong2024adaptiverag,
  title     = {Adaptive-{RAG}: Learning to Adapt Retrieval-Augmented Large Language Models through Question Complexity},
  author    = {Jeong, Soyeong and Baek, Jinheon and Cho, Sukmin and Hwang, Sung Ju and Park, Jong C.},
  booktitle = {Proceedings of NAACL},
  year      = {2024},
  doi       = {10.18653/v1/2024.naacl-long.389}
}

@misc{edge2024graphrag,
  title         = {From Local to Global: A Graph {RAG} Approach to Query-Focused Summarization},
  author        = {Edge, Darren and Trinh, Ha and Cheng, Newman and Bradley, Joshua and Chao, Alex and Mody, Apurva and Truitt, Steven and Metropolitansky, Dasha and Ness, Robert Osazuwa and Larson, Jonathan},
  year          = {2024},
  eprint        = {2404.16130},
  archivePrefix = {arXiv},
  doi           = {10.48550/arXiv.2404.16130}
}

@inproceedings{he2024gretriever,
  title     = {{G-Retriever}: Retrieval-Augmented Generation for Textual Graph Understanding and Question Answering},
  author    = {He, Xiaoxin and Tian, Yijun and Sun, Yifei and Chawla, Nitesh V. and Laurent, Thomas and LeCun, Yann and Bresson, Xavier and Hooi, Bryan},
  booktitle = {Advances in Neural Information Processing Systems},
  volume    = {37},
  year      = {2024},
  doi       = {10.52202/079017-4224}
}

@inproceedings{sun2024thinkongraph,
  title     = {Think-on-Graph: Deep and Responsible Reasoning of Large Language Model on Knowledge Graph},
  author    = {Sun, Jiashuo and Xu, Chengjin and Tang, Lumingyuan and Wang, Saizhuo and Lin, Chen and Gong, Yeyun and Ni, Lionel and Shum, Heung-Yeung and Guo, Jian},
  booktitle = {International Conference on Learning Representations},
  year      = {2024},
  url       = {https://openreview.net/forum?id=nnVO1PvbTv}
}

@inproceedings{mavromatis2025gnnrag,
  title     = {{GNN-RAG}: Graph Neural Retrieval for Efficient Large Language Model Reasoning on Knowledge Graphs},
  author    = {Mavromatis, Costas and Karypis, George},
  booktitle = {Findings of ACL},
  pages     = {16682--16699},
  year      = {2025},
  doi       = {10.18653/v1/2025.findings-acl.856}
}

@inproceedings{shen2025gear,
  title     = {{GeAR}: Graph-Enhanced Agent for Retrieval-Augmented Generation},
  author    = {Shen, Zhili and Diao, Chenxin and Vougiouklis, Pavlos and Merita, Pascual and Piramanayagam, Shriram and Chen, Enting and Graux, Damien and Melo, Andre and Lai, Ruofei and Jiang, Zeren and Li, Zhongyang and Qi, Ye and Ren, Yang and Tu, Dandan and Pan, Jeff Z.},
  booktitle = {Findings of ACL},
  pages     = {12049--12072},
  year      = {2025},
  doi       = {10.18653/v1/2025.findings-acl.624}
}

@inproceedings{lee2025hybgrag,
  title     = {{HybGRAG}: Hybrid Retrieval-Augmented Generation on Textual and Relational Knowledge Bases},
  author    = {Lee, Meng-Chieh and Zhu, Qi and Mavromatis, Costas and Han, Zhen and Adeshina, Soji and Ioannidis, Vassilis N. and Rangwala, Huzefa and Faloutsos, Christos},
  booktitle = {Proceedings of ACL},
  pages     = {879--893},
  year      = {2025},
  doi       = {10.18653/v1/2025.acl-long.43}
}

@misc{luo2025graphr1,
  title         = {{Graph-R1}: Towards Agentic {GraphRAG} Framework via End-to-End Reinforcement Learning},
  author        = {Luo, Haoran and E, Haihong and Chen, Guanting and Lin, Qika and Guo, Yikai and Xu, Fangzhi and Kuang, Zemin and Song, Meina and Wu, Xiaobao and Zhu, Yifan and Tuan, Luu Anh},
  year          = {2025},
  eprint        = {2507.21892},
  archivePrefix = {arXiv},
  doi           = {10.48550/arXiv.2507.21892}
}

@article{peng2025graphragsurvey,
  title   = {Graph Retrieval-Augmented Generation: A Survey},
  author  = {Peng, Boci and Zhu, Yun and Liu, Yongchao and Bo, Xiaohe and Shi, Haizhou and Hong, Chuntao and Zhang, Yan and Tang, Siliang},
  journal = {ACM Transactions on Information Systems},
  volume  = {44},
  number  = {2},
  articleno = {35},
  year    = {2026},
  doi     = {10.1145/3777378}
}

@misc{singh2025agenticragsurvey,
  title         = {Agentic Retrieval-Augmented Generation: A Survey on Agentic {RAG}},
  author        = {Singh, Aditi and Ehtesham, Abul and Kumar, Saket and Khoei, Tala Talaei and Vasilakos, Athanasios V.},
  year          = {2025},
  eprint        = {2501.09136},
  archivePrefix = {arXiv},
  doi           = {10.48550/arXiv.2501.09136}
}

@article{liu2026graphagents,
  title   = {Graph-Augmented Large Language Model Agents},
  author  = {Liu, Yixin and Zhang, Guibin and Wang, Kun and Li, Shiyuan and Pan, Shirui},
  journal = {IEEE Intelligent Systems},
  volume  = {41},
  number  = {2},
  pages   = {45--55},
  year    = {2026},
  doi     = {10.1109/MIS.2025.3642667}
}

@inproceedings{xiang2026whengraphs,
  title     = {When to Use Graphs in {RAG}: A Comprehensive Analysis for Graph Retrieval-Augmented Generation},
  author    = {Xiang, Zhishang and Wu, Chuanjie and Zhang, Qinggang and Chen, Shengyuan and Hong, Zijin and Huang, Xiao and Su, Jinsong},
  booktitle = {International Conference on Learning Representations},
  year      = {2026},
  eprint    = {2506.05690},
  url       = {https://arxiv.org/abs/2506.05690}
}

@inproceedings{ferrazzi2026agenticworth,
  title     = {Is Agentic {RAG} Worth It? An Experimental Comparison of {RAG} Approaches},
  author    = {Ferrazzi, Pietro and Cvjeti{\'c}anin, Milica and Piraccini, Alessio and Giannuzzi, Davide},
  booktitle = {Proceedings of ACL: Industry Track},
  pages     = {56--75},
  year      = {2026},
  doi       = {10.18653/v1/2026.acl-industry.5}
}

@inproceedings{gao2023alce,
  title     = {Enabling Large Language Models to Generate Text with Citations},
  author    = {Gao, Tianyu and Yen, Howard and Yu, Jiatong and Chen, Danqi},
  booktitle = {Proceedings of EMNLP},
  pages     = {6465--6488},
  year      = {2023},
  doi       = {10.18653/v1/2023.emnlp-main.398}
}

@inproceedings{es2024ragas,
  title     = {{RAGAS}: Automated Evaluation of Retrieval Augmented Generation},
  author    = {Es, Shahul and James, Jithin and Espinosa-Anke, Luis and Schockaert, Steven},
  booktitle = {Proceedings of EACL: System Demonstrations},
  pages     = {150--158},
  year      = {2024},
  doi       = {10.18653/v1/2024.eacl-demo.16}
}

@inproceedings{saadfalcon2024ares,
  title     = {{ARES}: An Automated Evaluation Framework for Retrieval-Augmented Generation Systems},
  author    = {Saad-Falcon, Jon and Khattab, Omar and Potts, Christopher and Zaharia, Matei},
  booktitle = {Proceedings of NAACL},
  pages     = {338--354},
  year      = {2024},
  doi       = {10.18653/v1/2024.naacl-long.20}
}

@inproceedings{vu2024freshllms,
  title     = {{FreshLLMs}: Refreshing Large Language Models with Search Engine Augmentation},
  author    = {Vu, Tu and others},
  booktitle = {Findings of ACL},
  pages     = {13697--13720},
  year      = {2024},
  doi       = {10.18653/v1/2024.findings-acl.813}
}

@inproceedings{qian2024timer4,
  title     = {{TimeR4}: Time-Aware Retrieval-Augmented Large Language Models for Temporal Knowledge Graph Question Answering},
  author    = {Qian, Xinying and Zhang, Ying and Zhao, Yu and Zhou, Baohang and Sui, Xuhui and Zhang, Li and Song, Kehui},
  booktitle = {Proceedings of EMNLP},
  pages     = {6942--6952},
  year      = {2024},
  doi       = {10.18653/v1/2024.emnlp-main.394}
}

@inproceedings{zou2025poisonedrag,
  title     = {{PoisonedRAG}: Knowledge Corruption Attacks to Retrieval-Augmented Generation of Large Language Models},
  author    = {Zou, Wei and Geng, Runpeng and Wang, Binghui and Jia, Jinyuan},
  booktitle = {USENIX Security Symposium},
  pages     = {3827--3844},
  year      = {2025},
  url       = {https://www.usenix.org/conference/usenixsecurity25/presentation/zou-poisonedrag}
}

@inproceedings{chen2024agentpoison,
  title     = {{AgentPoison}: Red-Teaming {LLM} Agents via Poisoning Memory or Knowledge Bases},
  author    = {Chen, Zhaorun and Xiang, Zhen and Xiao, Chaowei and Song, Dawn and Li, Bo},
  booktitle = {Advances in Neural Information Processing Systems},
  volume    = {37},
  year      = {2024},
  doi       = {10.52202/079017-4136}
}

@article{zhao2026knowledgepoisoning,
  title   = {Exploring Knowledge Poisoning Attacks to Retrieval-Augmented Generation},
  author  = {Zhao, Tianzhe and Chen, Jiaoyan and Ru, Yanchi and Zhu, Haiping and Hu, Nan and Liu, Jun and Lin, Qika},
  journal = {Information Fusion},
  volume  = {127},
  pages   = {103900},
  year    = {2026},
  doi     = {10.1016/j.inffus.2025.103900}
}

@inproceedings{liang2026graphragfire,
  title     = {{GraphRAG} under Fire},
  author    = {Liang, Jiacheng and Wang, Yuhui and Li, Changjiang and Jiang, Tanqiu and Zhu, Rongyi and Gong, Neil and Wang, Ting},
  booktitle = {IEEE Symposium on Security and Privacy},
  pages     = {195--212},
  year      = {2026},
  doi       = {10.1109/SP63933.2026.00070}
}

@inproceedings{zhan2024injecagent,
  title     = {{InjecAgent}: Benchmarking Indirect Prompt Injections in Tool-Integrated Large Language Model Agents},
  author    = {Zhan, Qiusi and Liang, Zhixiang and Ying, Zifan and Kang, Daniel},
  booktitle = {Findings of ACL},
  pages     = {10471--10506},
  year      = {2024},
  doi       = {10.18653/v1/2024.findings-acl.624}
}

@inproceedings{debenedetti2024agentdojo,
  title     = {{AgentDojo}: A Dynamic Environment to Evaluate Prompt Injection Attacks and Defenses for {LLM} Agents},
  author    = {Debenedetti, Edoardo and Zhang, Jie and Balunovi{\'c}, Mislav and Beurer-Kellner, Luca and Fischer, Marc and Tram{\`e}r, Florian},
  booktitle = {Advances in Neural Information Processing Systems},
  volume    = {37},
  year      = {2024},
  doi       = {10.52202/079017-2636}
}

@inproceedings{zeng2024ragprivacy,
  title     = {The Good and The Bad: Exploring Privacy Issues in Retrieval-Augmented Generation ({RAG})},
  author    = {Zeng, Shenglai and others},
  booktitle = {Findings of ACL},
  pages     = {4505--4524},
  year      = {2024},
  doi       = {10.18653/v1/2024.findings-acl.267}
}

@inproceedings{wu2025ragfairness,
  title     = {Does {RAG} Introduce Unfairness in {LLM}s? Evaluating Fairness in Retrieval-Augmented Generation Systems},
  author    = {Wu, Xuyang and Li, Shuowei and Wu, Hsin-Tai and Tao, Zhiqiang and Fang, Yi},
  booktitle = {Proceedings of COLING},
  pages     = {10021--10036},
  year      = {2025},
  url       = {https://aclanthology.org/2025.coling-main.669/}
}

@inproceedings{cemri2025masfailures,
  title     = {Why Do Multi-Agent {LLM} Systems Fail?},
  author    = {Cemri, Mert and others},
  booktitle = {Advances in Neural Information Processing Systems: Datasets and Benchmarks Track},
  year      = {2025},
  doi       = {10.52202/085713-4082}
}

@inproceedings{shen2025propagation,
  title     = {Understanding the Information Propagation Effects of Communication Topologies in {LLM}-Based Multi-Agent Systems},
  author    = {Shen, Xu and others},
  booktitle = {Proceedings of EMNLP},
  pages     = {12347--12361},
  year      = {2025},
  doi       = {10.18653/v1/2025.emnlp-main.623}
}

@inproceedings{south2025delegation,
  title     = {Position: {AI} Agents Need Authenticated Delegation},
  author    = {South, Tobin and Marro, Samuele and Hardjono, Thomas and Mahari, Robert and Whitney, Cedric Deslandes and Chan, Alan and Pentland, Alex},
  booktitle = {Proceedings of the 42nd International Conference on Machine Learning},
  series    = {Proceedings of Machine Learning Research},
  volume    = {267},
  pages     = {82211--82231},
  year      = {2025},
  url       = {https://proceedings.mlr.press/v267/south25a.html}
}

@article{green2019algorithmloop,
  title   = {The Principles and Limits of Algorithm-in-the-Loop Decision Making},
  author  = {Green, Ben and Chen, Yiling},
  journal = {Proceedings of the ACM on Human-Computer Interaction},
  volume  = {3},
  number  = {CSCW},
  pages   = {1--24},
  year    = {2019},
  doi     = {10.1145/3359152}
}

@article{lyons2021contestability,
  title   = {Conceptualising Contestability: Perspectives on Contesting Algorithmic Decisions},
  author  = {Lyons, Henrietta and Velloso, Eduardo and Miller, Tim},
  journal = {Proceedings of the ACM on Human-Computer Interaction},
  volume  = {5},
  number  = {CSCW1},
  pages   = {1--25},
  year    = {2021},
  doi     = {10.1145/3449180}
}

@inproceedings{raji2020auditing,
  title     = {Closing the {AI} Accountability Gap: Defining an End-to-End Framework for Internal Algorithmic Auditing},
  author    = {Raji, Inioluwa Deborah and Smart, Andrew and White, Rebecca N. and Mitchell, Margaret and Gebru, Timnit and Hutchinson, Ben and Smith-Loud, Jamila and Theron, Daniel and Barnes, Parker},
  booktitle = {Proceedings of the Conference on Fairness, Accountability, and Transparency},
  pages     = {33--44},
  year      = {2020},
  doi       = {10.1145/3351095.3372873}
}

@techreport{tabassi2023airmf,
  title       = {Artificial Intelligence Risk Management Framework ({AI RMF} 1.0)},
  author      = {Tabassi, Elham},
  institution = {National Institute of Standards and Technology},
  number      = {NIST AI 100-1},
  year        = {2023},
  doi         = {10.6028/NIST.AI.100-1}
}

@book{worldhealthorganization2021ethics,
  title     = {Ethics and Governance of Artificial Intelligence for Health},
  author    = {{World Health Organization}},
  publisher = {World Health Organization},
  year      = {2021},
  isbn      = {978-92-4-002920-0}
}

@article{tomasev2020aiforsocialgood,
  title   = {{AI} for Social Good: Unlocking the Opportunity for Positive Impact},
  author  = {Tomasev, Nenad and others},
  journal = {Nature Communications},
  volume  = {11},
  pages   = {2468},
  year    = {2020},
  doi     = {10.1038/s41467-020-15871-z}
}

\end{document}